\documentclass{article} 
\usepackage{iclr2020_conference,times}

\usepackage{amsmath,amsfonts,bm}

\def\eqref#1{equation~\ref{#1}}

\def\1{\bm{1}}

\DeclareMathAlphabet{\mathsfit}{\encodingdefault}{\sfdefault}{m}{sl}
\SetMathAlphabet{\mathsfit}{bold}{\encodingdefault}{\sfdefault}{bx}{n}

\usepackage{hyperref}
\usepackage{url}

\usepackage{amsmath}  
\usepackage{graphicx}
\usepackage{algorithm}
\usepackage{algorithmic}
\usepackage{diagbox}
\usepackage{subcaption}
\usepackage{authblk}

\title{Improving One-shot NAS by Suppressing the Posterior Fading}

\makeatletter
\newcommand{\printfnsymbol}[1]{%
  \textsuperscript{\@fnsymbol{#1}}%
}
\makeatother
\author[1]{Xiang Li\thanks{Equal Contribution}\ \ } 
\author[2]{Chen Lin\printfnsymbol{1}}
\author[2]{Chuming Li}
\author[2]{Ming Sun}
\author[2]{Wei Wu}
\author[2]{Junjie Yan}
\author[3]{Wanli Ouyang}
\affil[1]{Brown University}
\affil[2]{Sensetime Group Limited}
\affil[3]{The University of Sydney}
\affil[ ]{\texttt {xiang\_li\_1@brown.edu; wanli.ouyang@sydney.edu.au;  \{linchen,lichuming,sunming1,wuwei,yanjunjie\}@sensetime.com}}

\iclrfinalcopy 
\begin{document}

\maketitle

\begin{abstract}
 There is a growing interest in automated neural architecture search (NAS). To improve the efficiency of NAS, previous approaches adopt \textit{weight sharing} method to force all models share the same set of weights. However, it has been observed that a model performing better with shared weights does not necessarily perform better when trained alone. In this paper, we analyse existing weight sharing  one-shot NAS approaches from a Bayesian point of view and identify the \textbf{posterior fading problem}, which compromises the effectiveness of shared weights. To alleviate this problem,  we present a practical approach to guide the parameter posterior towards its true distribution. Moreover, a hard latency constraint is introduced during the search so that the desired latency can be achieved. The resulted method, namely Posterior Convergent NAS (PC-NAS), achieves state-of-the-art performance under standard GPU latency constraint on ImageNet.
In our small search space, our model PC-NAS-S attains $76.8 \%$ top-1 accuracy, $2.1\%$ higher than MobileNetV2 (1.4x) with the same latency. When adopted to the large search space, PC-NAS-L achieves $78.1 \%$ top-1 accuracy within 11ms. The discovered architecture also transfers well to other computer vision applications such as object detection and person re-identification.
\end{abstract}
\section{Introduction}

Neural network design requires extensive experiments by human experts. In recent years,  
neural architecture search \citep{zoph2016neural, liu2018progressive, zhong2018practical, liautoloss, linautoaugmentation} has emerged as a promising tool to alleviate the cost of human efforts on manually balancing accuracy and resources constraint. 

Early works of NAS \citep{real2018regularized, Thomashillclimbing} achieve promising results but have to resort to search only using proxy or subsampled dataset due to its large computation expense.
Recently, the attention is drawn to improve the search efficiency via sharing weights across models \citep{Leoneshot,  enas}. 
Generally, weight sharing approaches utilize an over-parameterized network (supergraph) containing every single model, which can be mainly divided into two categories. 

The first category is continuous relaxation method \citep{liu2018darts, cai2018proxylessnas}, which keeps a set of so called architecture parameters to represent the model, and updates these parameters alternatively with supergraph weights. The resulted model is obtained using the architecture parameters at convergence. 
The continuous relaxation method entails the rich-get-richer problem \citep{Georegeunderstandingnas}, which means that a better-performed model at the early stage would be trained more frequently (or have larger learning rates). This introduces bias and instability to the search process. 

Another category is referred to as one-shot method \citep{Brockoneshot, Zichao, Leoneshot, xiangxiangfairnas}, which divides the NAS proceedure into a training stage and a searching stage. 
In the training stage, the supergraph is optimized along with either dropping out each operator with certain probability or sampling uniformly among candidate architectures.
In the search stage, a search algorithm is applied to find the architecture with the highest validation accuracy with shared weights. 
The one-shot approach ensures the fairness among all models by sampling architecture or dropping out operator uniformly. However, as identified in \citep{Georegeunderstandingnas, xiangxiangfairnas, Leoneshot}, the validation accuracy of the model with shared weights is not predictive to its true performance.

In this paper, we formulate NAS as a Bayesian model selection problem \citep{hugebayesian}. With this formulation, we can obtain a comprehensive understanding of one-shot approaches. We show that shared weights are actually a maximum likelihood estimation of a proxy distribution to the true parameter distribution. Further, we identify the common issue of weight sharing, which we call \textbf{Posterior Fading}, i.e., the KL-divergence between true parameter posterior and proxy posterior also increases with the number of models contained in the supergraph. 

To alleviate the aforementioned problem, we proposed a practical approach to guide the convergence of the proxy distribution towards the true parameter posterior. Specifically, our approach divides the training of supergraph into several intervals.
We maintain a pool of high potential partial models and progressively update this pool after each interval . At each training step, a partial model is sampled from the pool and complemented to a full model. To update the partial model pool, we generate candidates by extending each partial model and evaluate their potentials, the top ones among which form the new pool size. Since the search space is shrinked in the upcoming training interval, the parameter posterior get close to the desired true posterior during this procedure.
Main contributions of our work is concluded as follows:
\begin{itemize}
\item We analyse the one-shot approaches from a Bayesian point of view and identify the associated disadvantage which we call Posterior Fading.


\item Inspired by the theoretical discovery, we introduce a novel NAS algorithm which guide the proxy distribution to converge towards the true parameter posterior. 
\end{itemize}

We apply our proposed approach to ImageNet classification \citep{Russakovskyimagenet} and achieve strong empirical results. In one typical search space \citep{cai2018proxylessnas}, our PC-NAS-S attains $76.8 \%$ top-1 accuracy, $0.5 \%$ higher and $20 \%$ faster than EfficientNet-B0 \citep{Leefficientnet}, which is the previous state-of-the-art model in mobile setting. To show the strength of our method, we apply our algorithm to a larger search space, our PC-NAS-L boosts the accuracy to $78.1 \%$. 

\section{Related work}
\label{sec:related_work}
Increasing interests are drawn to automating the design of neural network with machine learning techniques such as reinforcement learning or neuro-evolution, which is usually referred to as neural architecture search(NAS) \citep{Miller, Liuhierarchical, real2017large, zoph2016neural, Baker, Lin, liu2018darts, cai2018proxylessnas}. This type of NAS is typically considered as an agent-based explore and exploit process, where an agent (e.g. an evolution mechanism or a recurrent neural network(RNN)) is introduced to explore a given architecture space with training a network in the inner loop to get an evaluation for guiding exploration. Such methods are computationally expensive and hard to be used on large-scale
datasets, e.g. ImageNet.

Recent works \citep{enas, Brocksmash, liu2018darts, cai2018proxylessnas} try to alleviate this
computation cost via modeling NAS as a single training process of an over-parameterized network that comprises all candidate models, in which weights of the same operators in different models are shared. ENAS \citep{enas} reduces the computation cost by orders of magnitude, while requires an RNN agent and focuses on small-scale datasets (e.g. CIFAR10). One-shot NAS \citep{Brockoneshot} trains the over-parameterized network along with droping out each operator with increasing probability. Then it use the pre-trained over-parameterized network to evaluate randomly sampled architectures. DARTS \citep{liu2018darts} additionally introduces a real-valued architecture parameter for each operator and alternately train operator weights and architecture parameters by back-propagation. ProxylessNAS \citep{cai2018proxylessnas} binarize the real-value parameters in DARTS to save the GPU cumputation and memory for training the over-parameterized network.

The paradigm of ProxylessNAS \citep{cai2018proxylessnas} and DARTS \citep{liu2018darts} introduce unavoidable bias since operators of models performing well in the beginning will easily get trained more and normally keep being better than other. But they are not necessarily superior than others when trained from scratch. 

Other relevant works are ASAP \citep{noy2019asap} and XNAS \citep{nayman2019xnas}, which introduce pruning during the training of over-parameterized networks to improve the efficiency of NAS. Similarly, we start with an over-parameterized network and then reduce the search space to derive the optimized architecture. The distinction is that they focus on the speed-up of training and only prune by evaluating the architecture parameters, while we improves the rankings of models and evaluate operators direct on validation set by the performance of models containing it.

\section{Methods}
\label{sec:methods}
In this section, we first formulate neural architecture search in a Bayesian manner. Utilizing this setup, we introduce our PC-NAS approach and analyse its advantage against previous approach. Finally, we discuss the search algorithm combined with latency constraint.




\subsection{A Probabilistic Setup for Model Uncertainty}
Suppose $K$ models $\mathcal{M} = \{\mathbf{m}_1, . . . , \mathbf{m}_K\}$ are under consideration for data $\mathcal{D}$, and $p(\mathcal{D} | \theta_k, \mathbf{m}_k)$ describes the probability density of data $\mathcal{D}$ given model $\mathbf{m}_k$ and its associated parameters $\theta_k$. The Bayesian approach proceeds by assigning a prior probability distribution $p(\theta_k |\mathbf{m}_k)$ to the parameters of each model, and a prior probability $p(\mathbf{m}_k)$ to each model. 

In order to ensure fairness among all models, we set the model prior $p(\mathbf{m}_k)$ a uniform distribution. Under previous setting, we can drive
\begin{equation}
\label{eqn:modelposterior}
p(\mathbf{m}_k|\mathcal{D}) = \
\frac{p(\mathcal{D}|\mathbf{m}_k)p(\mathbf{m}_k)}{\sum_kp(\mathcal{D}|\mathbf{m}_k)p(\mathbf{m}_k)},
\end{equation}
where
\begin{equation}
\label{eqn:modelposteriorwhere}
p(\mathcal{D}|\mathbf{m}_k) = \
\int\! p(\mathcal{D} | \theta_k, \mathbf{m}_k) p(\theta_k|\mathbf{m}_k)\mathrm{d}\theta_k.
\end{equation}
Since $p(\mathbf{m}_k)$ is uniform, the Maximum Likelihood Estimation (MLE) of $\mathbf{m}_k$ is just the maximum of (\ref{eqn:modelposteriorwhere}). It can be inferred that, $p(\theta_k|\mathbf{m}_k)$ is crucial to the solution of the model selection. We are interested in attaining the model with highest test accuracy in a trained alone manner, thus the parameter prior $p(\theta_k|\mathbf{m}_k)$ is just the posterior $p_{\text{alone}}(\theta_k|\mathbf{m}_k,\mathcal{D})$ which means the distribution of $\theta_k$ when $\mathbf{m}_k$ is trained alone on dataset $\mathcal{D}$. Thus we would use the term true parameter posterior to refer $p_{\text{alone}}(\theta_k|\mathbf{m}_k,\mathcal{D})$.

\begin{figure}[t]
    \centering
    \includegraphics[width=0.8\linewidth]{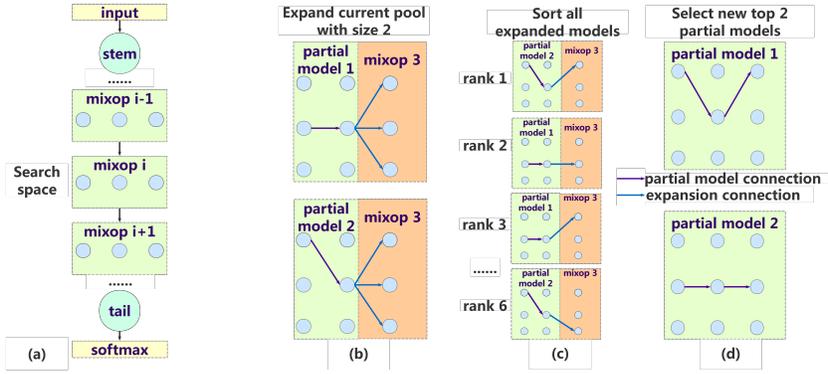}
    \caption{One example of search space(a) and PC-NAS process(b)(c)(d). Each mixed opperator consists of $N$(=3 in this figure) operators. However, only one operator in each mixop is invoked at a time for each batch.
    In (b), partial models 1 and 2 in the pool consist of choices in mixop 1 and 2.
    We extend these 2 partial models to one mixop 3. 6 extended candidate models are evaluated and ranked in(c). In (d), the new pool consists of the top-2 candidate models ranked in (c).}
    \label{fig:figure1}
\end{figure}

\subsection{Network Architecture Selection In a Bayesian Point of View}
We constrain our discussion on the setting which is frequently used in NAS literature.
As a building block of our search space, a mixed operator (mixop), denoted by $\mathbb{O}=\{O_1\ldots, O_N\}$,   contains $N$ different choices of candidate operators $O_i$ for $i=1, \ldots N$ in parallel. The search space is defined by $L$ mixed operators (layers) connected sequentially interleaved by downsampling as in Fig.~\ref{fig:figure1}(a). 
The network architecture (model) $\mathbf{m}$ is defined by a vector $[o_1, o_2, ..., o_L]$, $o_l\in \mathbb{O}$ representing the choice of operator for layer $l$. The parameter for the operator $o$ at the $l$-th layer is denoted as $\theta_{lo}$. The parameters of the supergraph are denoted by $\theta$ which includes $\{\theta_{lo} |l\in\{1,2,...,L\}, o\in\mathbb{O}\}$.
In this setting, the parameters of each candidate operator are shared among multiple architectures. The parameters related with a specific model $\mathbf{m}_k$ is denoted as $\theta_{\mathbf{m}_k}=\theta_{1,o_1},\theta_{2,o_2}, ..., \theta_{L,o_L}$, which is a subset of the parameters of the supergraph $\theta$, the rest of the parameters are denoted as $\bar\theta_{\mathbf{m_k}}$, \textit{i.e.} $\theta_{\mathbf{m}_k}\cap  \bar\theta_{\mathbf{m_k}} =\emptyset, \theta_{\mathbf{m}_k}\cup  \bar\theta_{\mathbf{m_k}} =\theta$. 
The posterior of all parameters $\theta$ given $\mathbf{m}_k$ has the property $p_{\text{alone}}(\theta_{\mathbf{m}_k}|\mathbf{m}_k, \mathcal{D}) =p_{\text{alone}}(\theta_{\mathbf{m}_k}|\mathbf{m}_k, \mathcal{D})p_{\text{alone}}(\bar{\theta}_k|\mathbf{m}_k, \mathcal{D})$. Implied by the fact that $\bar{\theta}_{\mathbf{m}_k}$ does not affect the prediction of $\mathbf{m}_k$ and also not updated during training, $p_{\text{alone}}(\bar{\theta}_{\mathbf{m}_k}|\mathbf{m}_k, \mathcal{D})$ is uniformly distributed, .
Obtaining the $p_{\text{alone}}(\theta_k|\mathbf{m}_k, \mathcal{D})$ or a MLE of it for each single model is computationally intractable. 
Therefore, the one-shot method trains the supergraph by dropping out each operator ~\citep{Brockoneshot} or sampling different architectures~\citep{Leoneshot,xiangxiangfairnas} and utilize the shared weights to evaluate single model. 
In this work, we adopt the latter training paradigm while the former one could be easily generalized. 
Suppose we sample a model $\mathbf{m}_k$ and optimize the supergraph with a mini-batch of data 
based on the objective function $L_{\text{alone}}$:
\begin{equation}
\label{eqn:oneshotpriorMLEsub}
-\log{p_{\text{alone}}(\theta|\mathbf{m}_k, \mathcal{D})} \propto  L_{\text{alone}}(\theta, \mathbf{m}_k, \mathcal{D}) =\
-\log{p_{\text{alone}}(\mathcal{D}|\theta, \mathbf{m}_k)} - \log{p(\theta|\mathbf{m}_k)},
\end{equation}
where $-\log{p(\theta|\mathbf{m}_k)}$ is a regularization term. Thus minimizing this objective equals to making MLE to $p_{\text{alone}}(\theta|\mathbf{m}_k, \mathcal{D})$. When training the supergraph, we sample many models $\mathbf{m}_k$, and then train the parameters for these models, which corresponds to a stochastic approximation of the following objective function:
\begin{equation}
\label{eqn:oneshotpriorMLE}
L_{\text{share}}(\theta, D) =\
\frac{1}{K}\sum_{k}L_{\text{alone}}(\theta, \mathbf{m}_k, \mathcal{D}).
\end{equation}
This is equivalent to adopting a proxy parameter posterior as follows:
\begin{equation}
\label{eqn:oneshotprior}
p_{\text{share}}(\theta| \mathcal{D}) = \
\frac{1}{Z}\prod_k p_{\text{alone}}(\theta|\mathbf{m}_k, \mathcal{D}), 
\end{equation}
\begin{equation}
\label{eqn:oneshotprior2}
-\log{p_{\text{share}}(\theta| \mathcal{D})} = \
-\sum_k \log{p_{\text{alone}}(\theta|\mathbf{m}_k, \mathcal{D}) } + \log{Z}.
\end{equation}
Maximizing $p_{\text{share}}(\theta|D)$ is equivalent to minimizing $L_{\text{share}}$.

We take one step further to assume that the parameters at each layer are independent, i.e.
\begin{equation}
\label{eqn:palonemeanfield}
p_{\text{alone}}(\theta|\mathbf{m}_k,\mathcal{D})=\
\prod_{l,o} p_{\text{alone}}(\theta_{l,o}|\mathbf{m}_k,\mathcal{D}).
\end{equation}
Due to the independence, we have
\begin{equation}
\label{eqn:psharemeanfield}
\begin{aligned}
p_{\text{share}}(\theta| \mathcal{D}) &= \prod_k \prod_{l,o} p_{\text{alone}}(\theta_{l,o}|\mathbf{m}_k, \mathcal{D})= \prod_{l,o} p_{\text{share}}(\theta_{l,o}| \mathcal{D}),
\end{aligned}
\end{equation}
where
\begin{equation}
\label{eqn:psharemeanfieldsecond}
p_{\text{share}}(\theta_{l,o}|\mathcal{D})=\
\prod_{k} p_{\text{alone}}(\theta_{l,o}|\mathbf{m}_k,\mathcal{D}).
\end{equation}
The KL-divergence between $p_{\text{alone}}(\theta_{l,o}|\mathbf{m}_k,\mathcal{D})$ and $p_{\text{share}}(\theta_{l,oj}|\mathcal{D})$ is as follows:
\begin{equation}
\label{eqn:KL}
\begin{aligned}
D_{\mathcal{KL}} \Big(p_{\text{alone}}(\theta_{l,o}|\mathbf{m}_k,\mathcal{D})  \  \Big|\Big|  \  p_{\text{share}}(\theta_{l,o}| \mathcal{D})\Big) &= \int p_{\text{alone}}(\theta_{l,o}|\mathbf{m}_k,\mathcal{D}) \log{\frac{p_{\text{alone}}(\theta_{l,o}|\mathbf{m}_k,\mathcal{D})}{p_{\text{share}}(\theta_{l,o}| \mathcal{D})}} \mathrm{d}\theta \\
&= \int p_{\text{alone}}(\theta_{l,o}|\mathbf{m}_k,\mathcal{D}) \log{\frac{p_{\text{alone}}(\theta_{l,o}|\mathbf{m}_k,\mathcal{D})}{\prod_i p_{\text{alone}}(\theta_{l,o}|\mathbf{m}_i, \mathcal{D})}} \mathrm{d}\theta \\
&= \sum_{i\neq k}-\int p_{\text{alone}}(\theta_{l,o}|\mathbf{m}_k,\mathcal{D}) \log{p_{\text{alone}}(\theta_{l,o}|\mathbf{m}_i, \mathcal{D})} \mathrm{d}\theta. \\
\end{aligned}
\end{equation}
Since the KL-divergence is just the summation of the cross-entropy of $p_{\text{alone}}(\theta_{l,o}|\mathbf{m}_k,\mathcal{D})$ and $p_{\text{alone}}(\theta_{l,o}|\mathbf{m}_i, \mathcal{D})$ where $i\neq k$. The cross-entropy term is always positive. Increasing the number of architectures would push $p_{\text{share}}$ away from $p_{\text{alone}}$, namely the \textbf{Posterior Fading}. We conclude that non-predictive problem originates naturally from one-shot supergraph training. 
Based on this analysis, if we effectively reduce the number of architectures in Eq.(\ref{eqn:KL}), the divergence would decrease, which motivates our design in the next section.

\begin{algorithm}
\caption{Potential: Evaluating the Potential of Partial Candidates}
\label{alo:eval}
\begin{algorithmic}
\STATE \textbf{Inputs:} $G$(supergraph), $L$(num of mixops in $G$), $\mathbf{m}'$(partial candidate), $Lat$(latency constraint), $S$(evaluation number), $D_{val}$ (validataion set)
\STATE Scores = $\emptyset$
\FOR{$i$ = 1 : $S$}
\STATE $\mathbf{m}^* = $ expand($\mathbf{m}'$) \quad \textit{randomly expand} $\mathbf{m}'$ \textit{to full depth} $L$
\IF{Latency($\mathbf{m}^*$) $>$ Lat} 
\STATE continue \quad \textit{dump samples that don't satisfy the latency constraint}
\ENDIF
\STATE acc = Acc($\mathbf{m}^*$, $D_{val}$) \quad \textit{inference} $\mathbf{m}^*$ \textit{for one batch and return its accuracy}
\STATE Scores.append(acc) \quad \textit{save accuracy}
\ENDFOR
\STATE \textbf{Outputs: }Average(Scores)
\end{algorithmic}
\end{algorithm}

\subsection{Posterior Convergent NAS}
One trivial way to mitigate the posterior fading problem is limit the number of candidate models inside the supergraph.
However, large number of candidate models is demanded for NAS to discover promising models.
Due to this conflict, we present PC-NAS which adopt progressive search space shrinking. 
The resulted algorithm divide the training of shared weights into $L$ intervals, where $L$ is the number of mixed operators in the search space. The number of training epochs of a single interval is denoted as $T_{i}.$

\textbf{Partial model pool} is a collection of partial models. At the $l$-th interval, a single partial model should contain $l-1$ selected operators $[o_1, o_2, ..., o_l]$. The size of partial model pool is denoted as $P$.  After the $l$-th interval, each partial model in the pool will be extended by the $N$ operators in $l$-th mixop. Thus there are $P\times N$ candidate extended partial models with length $l$. These candidate partial models are evaluated and the top-$P$ among which are used as the partial model pool for the interval $l+1$. An illustrative exmaple of partial model pool update is in Fig.~\ref{fig:figure1}(b)(c)(d).

\textbf{Candidate evaluation with latency constraint:} We simply define the potential of a partial model to be the expected validation accuracy of the models which contain the partial model. 
\begin{equation}
\label{eqn:partialpotential}
\text{Potential}({o_1, o_2, ..., o_l})=E_{ \mathbf{m} \in \left\{\mathbf{m}| m_i =o_i , \forall i \leq l \right\}}(Acc(\mathbf{m})).
\end{equation}
where the validation accuracy of model $\mathbf{m}$ is denoted by $Acc(\mathbf{m})$.
We estimate this value by uniformly sampling valid models and computing the average of their validation accuracy using one mini-batch. We use $S$ to denote the evaluation number, which is the total number of sampled models. We observe that when $S$ is large enough, the potential of a partial model is fairly stable and discriminative among candidates. See Algorithm.~\ref{alo:eval} for pseudo code. The latency constraint is imposed by dumping invalid full models when calculating potentials of extended candidates of partial models in the pool.

\textbf{Training based on partial model pool} The training iteration of the supergraph along with the partial model pool has two steps. 
First, for a partial model from the pool, we randomly sample the missing operator $\{o_{l+1}, o_{l+2}, ..., o_{L}\}$ to complement the partial model to a full model. Then we optimize $\theta$ using the sampled full model and mini-batch data. 
We
Initially, the partial model pool is empty. Thus the supergraph is trained by uniformly sampled models, which is identical to previous one-shot training stage. After the initial training, all operators in the first mixop are evaluated. The top $P$ operators forms the partial model pool in the second training stage. Then, the supergraph resume training and the training procedure is identical to the one discussed in last paragraph.
Inspired by warm-up, the first stage is set much more epochs than following stages denoted as $T_w$. The whole PC-NAS process is elaborated in algorithm.~\ref{alo:BNAS} The number of models in the shrinked search space at the interval $l$ is strictly less than interval $l-1$. At the final interval, the number of cross-entropy terms in Eq.(\ref{eqn:KL}) are P-1 for each architectures in final pool. Thus the parameter posterior of PC-NAS would move towards the true posterior during these intervals.
\begin{algorithm}
\caption{PC-NAS: Posterior Convergent Architecture Search}
\label{alo:BNAS}
\begin{algorithmic}
\STATE  \textbf{Inputs:} $P$(size of partial model pool), $G$(supergraph), $O_i$ (the ith operator in mixed operator), $L$(num of mixed operators in $G$), $T_w$(warm-up epochs), $T_i$(interval between updation of partial model pool), $D_{train}$(train set), $D_{val}$ (validataion set),  $Lat$(latency constraint)
\STATE  PartialModels = $\emptyset$
\STATE Warm-up($G$, $D_{train}$, $T_w$) \quad \textit{uniformly sample models from G and train}
\FOR{$I$ = 0:($L \cdot T_i -$1)}
\IF{$I$ mod $T_i$ == 0}
\STATE ExtendedPartialModels = $\emptyset$
\IF{PartialModels == $\emptyset$} 
\STATE ExtendedPartialModels.append($[O_i]$) \quad \textit{add all operator in the first mixop}
\ENDIF
\FOR{$\mathbf{m}$ in PartialModels}
\STATE ExtendedPartialModels.append(Extend($\mathbf{m}$,$ O_1$), ..., Extend($\mathbf{m}$,$O_N$)) 
\ENDFOR
\FOR{$\mathbf{m}'$ in ExtendedPartialModels}
\STATE $\mathbf{m}'$.potential = Potential($\mathbf{m}'$, $D_{val}$, $Lat$, $S$)  \quad \textit{evaluate the extended partial model}
\ENDFOR
\STATE PartialModels = Top(ExtendedPartialModels, $P$) \quad \textit{keep P best partial models}
\ENDIF
\STATE Train(PartialModels, $D_{train}$) \quad \textit{train one epoch using partial models} 
\ENDFOR
\STATE \textbf{Outputs:} PartialModels
\end{algorithmic}
\end{algorithm}
\section{Experiments Results}
We demonstrate the effectiveness of our methods on ImageNet, a large scale benchmark dataset, which contains 1,000,000 training samples with 1000 classes.
For this task, we focus on models that have high accuracy under certain GPU latency constraint. We search models using PC-NAS,
which progressively updates a partial model pool and trains shared weights. Then, we select the model with the highest potential in the pool and report its performance on the test set after training from scratch. Finally, we investigate the transferability of the model learned on ImageNet by evaluating it on two tasks, object detection and person re-identification.
\subsection{Training Details} 
\textbf{Dataset and latency measurement:} As a common practice, we randomly sample 50,000 images from the train set to form a validation set during the model search. We conduct our PC-NAS on the remaining images in train set. The original validation set is used as test set to report the performance of the model generated by our method. The latency is evaluated on Nvidia GTX 1080Ti and the batch size is set 16 to fully utilize GPU resources.

\textbf{Search spaces:} We use two search spaces. We benchmark our \textbf{small space} similar to ProxylessNAS \citep{cai2018proxylessnas} and FBNet \citep{wu2018fbnet} for fair comparison. To test our PC-NAS method in a more complicated search space, we add 3 more kinds of operators to the small space's mixoperators to construct our \textbf{large space}. Details of the two spaces are in ~\ref{search_space}.

\textbf{PC-NAS hyperparameters:} We use PC-NAS to search in both small and large space. To balance training time and performance, we set evaluation number $S = 900$ and partial model pool size $P = 5$ in both experiments. Ablation study of the two values is in ~\ref{sec:ab}. When updating weights of the supergraph, we adopt mini-batch nesterov SGD optimizer with momentum 0.9, cosine learning rate decay from 0.1 to 5e-4 and batch size 512, and L2 regularization with weight 1e-4. The warm-up epochs $T_w$ and shrinking interval $T_i$ are set 100 and 5, thus the total training of supergraph lasts $100+20\times 5=200$ epochs. After searching, we select the best one from the top 5 final partial models and train it from scratch. The hyperparameters used to train this best model are the same as that of supergraph and the training takes 300 epochs. We add squeeze-and-excitation layer to this model at the end of each operator and use mixup during the training of resulted model. 

\subsection{ImageNet Results}

\begin{table*}[ht]
        \caption{PC-NAS' Imagenet results compared with state-of-the-art methods in the \textit{mobile} setting. }
		\label{bigtable}
		\begin{center}
			\begin{tabular}{l|c|c|c|c}
				\hline
				\multicolumn{1}{c|}{\bf model}  &\multicolumn{1}{c|}{\bf space} &\multicolumn{1}{c|}{\bf params} &\multicolumn{1}{c|}{\bf latency(gpu)}
				&\multicolumn{1}{c}{\bf top-1 acc}\\ 
				\hline
				MobileNetV2 1.4x \citep{sandler2018mobilenetv2}	&-	& 6.9M 	&10ms	&74.7\% 			\\
				AmoebaNet-A\citep{real2018regularized} 		&-	& 5.1M	&23ms	&74.5\%			\\
				PNASNet \citep{liu2018progressive}			&5.6x$10^{14}$	& 5.1M	&25ms 	& 74.2\% 			\\
				MnasNet\citep{tan2018mnasnet}				&-	& 4.4M	&11ms	&74.8\% 			\\ 
				FBNet-C\citep{wu2018fbnet}				&$10^{21}$&5.5M	&-		&74.9\%			\\
				ProxylessNAS-gpu\citep{cai2018proxylessnas}     &$7^{21}$     	& 7.1M &8ms	&75.1\%			\\
				EfficientNet-B0\citep{Leefficientnet}				&-	& 5.3M	&13 ms	&76.3\% 			\\
				MixNet-S\citep{Tanmix}				&-&4.1M		&13 ms		&75.8\%			\\
				\hline
				PC-NAS-S                  &$10^{21}$&5.1M &10 ms &76.8\% \\
				PC-NAS-L                  &$20^{21}$&15.3M &11 ms &78.1\% \\
                \hline
				
			\end{tabular}
		\end{center}
\end{table*}

Table \ref{bigtable} shows the performance of our model on ImageNet. We set our target latency at $10 ms$ according to our measurement of mobile setting models on GPU.
Our search result in the small space, namely PC-NAS-S, achieves 76.8\% top-1 accuracy under our latency constraint, which is $0.5 \%$ higher than EffcientNet-B0 (in terms of absolute accuracy improvement), $1 \%$ higher than MixNet-S. 
If we slightly relax the time constraint, our search result from the large space, namly PC-NAS-L, achieves $78.1 \%$ top-1 accuracy, which improves top-1 accuracy by $1.8 \%$ compared to EfficientNet-B0, $2.3 \%$ compared to MixNet-S. Both PC-NAS-S and PC-NAS-L are faster than EffcientNet-b0 and MixNet-S. 
\begin{table}[ht]
\caption{Performance Comparison on COCO and Market-1501}
\label{cocomap}
\begin{center}
\begin{tabular}{c|c|c|c|c}
\hline
\multicolumn{1}{c|}{\bf backbone}  &\multicolumn{1}{c|}{\bf params} &\multicolumn{1}{c|}{\bf latency} &\multicolumn{1}{c|}{\bf COCO mAP } &\multicolumn{1}{c}{\bf Market-1501 mAP}\\ 
\hline
MobileNetV2      &3.5M	   &7ms    &31.7   &76.8\\
ResNet50         &25.5M    &15ms   &36.8   &80.9\\
ResNet101        &44.4M    &26ms   &39.4   &82.1\\
\hline
PC-NAS-L           &15.3M    &11ms   &38.5   &81.0\\
\hline
\end{tabular}
\end{center}
\end{table}
\subsection{Transferability of PC-NAS}
We validate our PC-NAS's transferability on two tasks, object detection and person re-identification.  We use COCO \citep{lin2014microsoft} dataset as benchmark for object detection and Market-1501 \citep{zheng2015scalable} for person re-identification. For the two dataset, PC-NAS-L pretrained on ImageNet is utilized as feature extractor, and is compared with other models under the same training script. For object detection, the experiment is conducted with the two-stage framework FPN \citep{fpn}. Table \ref{cocomap} shows the performance of our PC-NAS model on COCO and Market-1501. For COCO, our approach significantly surpasses the mAP of MobileNetV2 as well as ResNet50. Compare to the standard ResNet101 backbone, our model achieves comparable mAP quality with almost $1/3$ parameters and $2.3\times$ faster speed. Similar  phenomena are found on Market-1501.
\subsection{Ablation Study}\label{sec:ab}
\begin{figure}
  \centering
  \begin{subfigure}[t]{0.45\textwidth}
    \includegraphics[width=\textwidth]{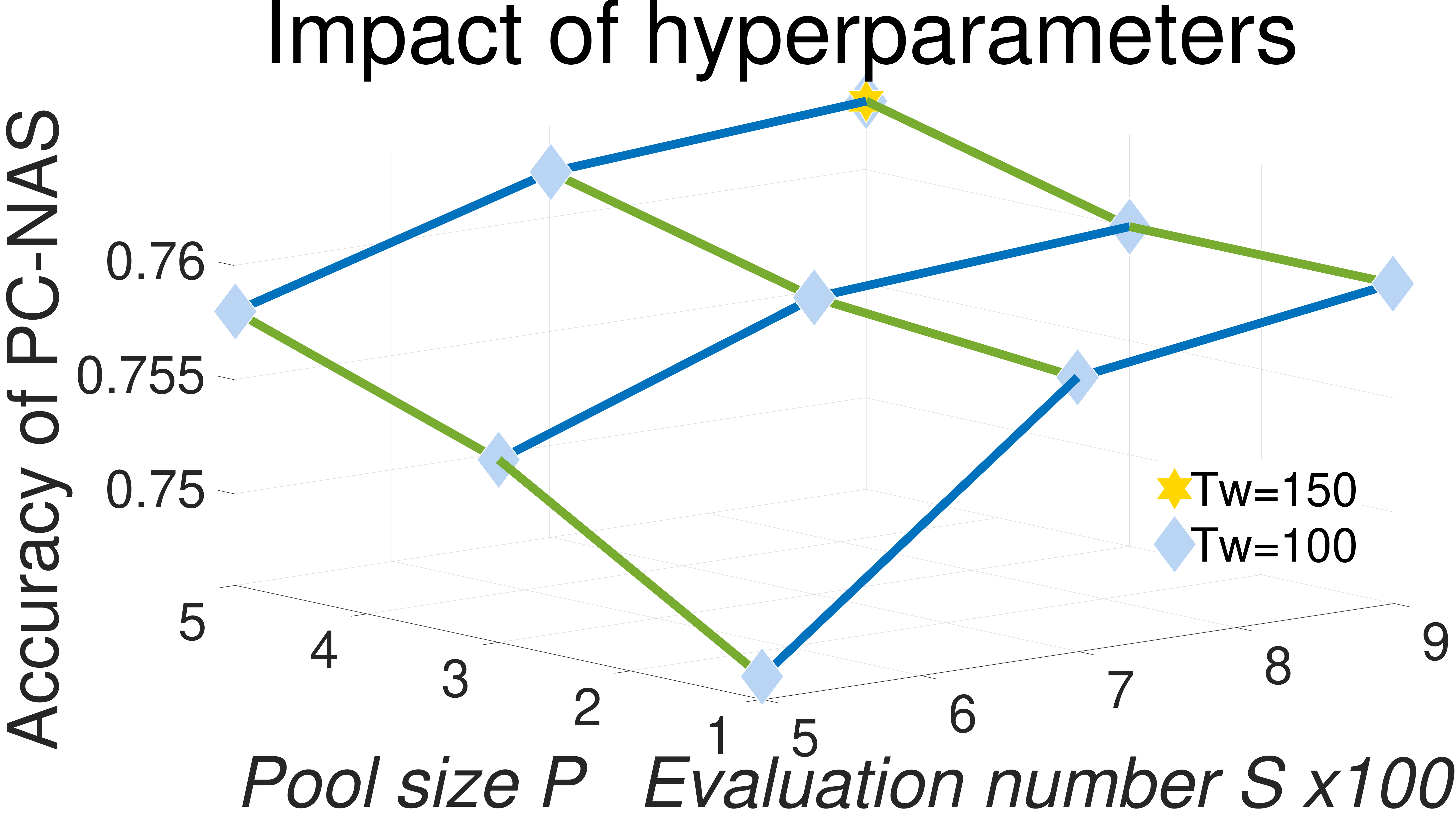}
    \caption{}
    \label{fig:hyper}
  \end{subfigure}
  \begin{subfigure}[t]{0.45\textwidth}
    \includegraphics[width=\textwidth]{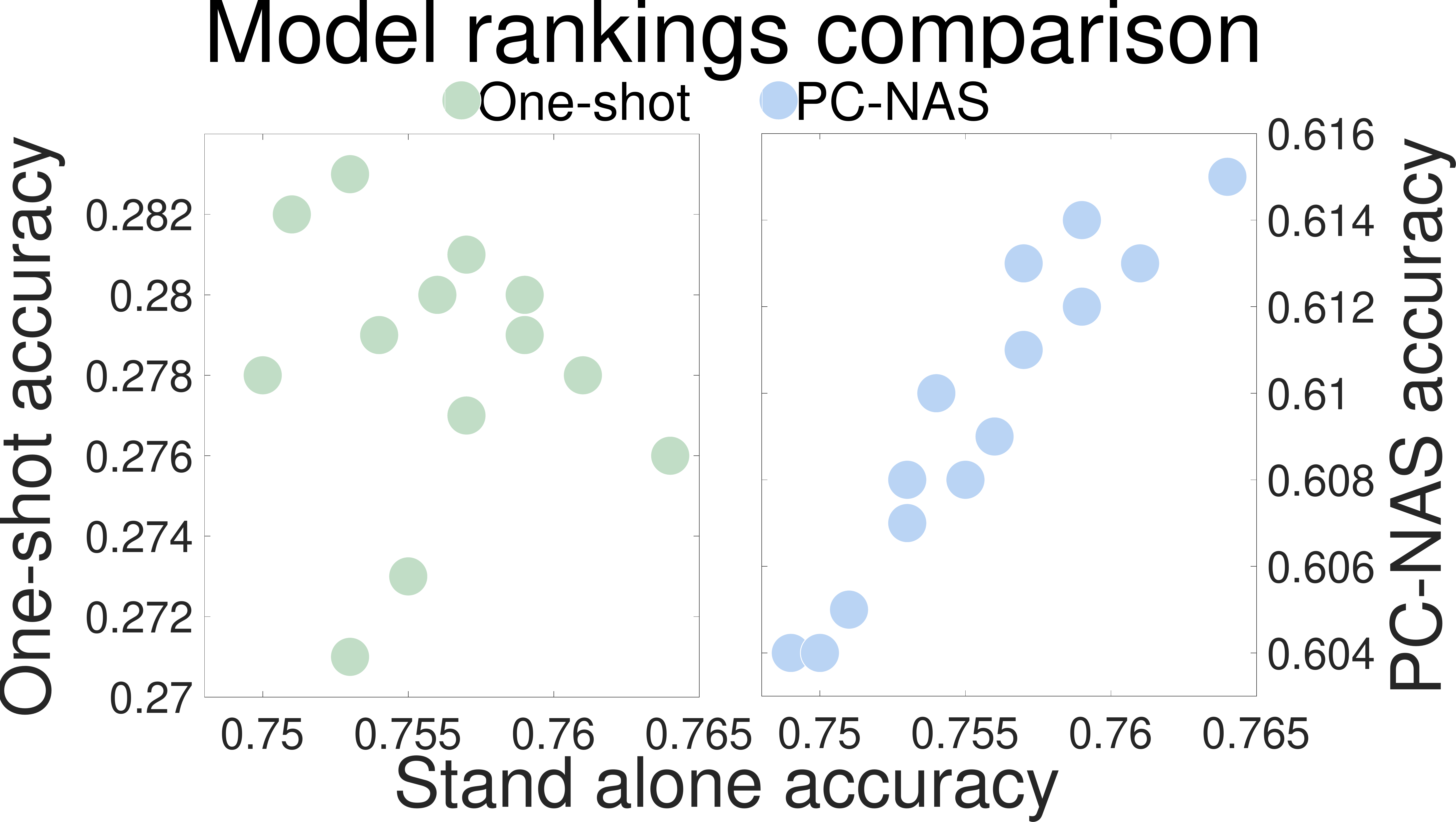}
    \caption{}
    \label{fig:correlation}
  \end{subfigure}
      \caption{(a): Influence of warm-up epochs $T_w$, partial model pool size $P$, and evaluation number $S$ to the resulted model. (b): Comparison of model rankings for One-Shot (left) and PC-NAS (right).}
\end{figure}

\textbf{Impact of hyperparameters:} In this section, we further study the impact of hyperparameters on our method within our small space on ImageNet. The hyperparameters include warm-up, training epochs Tw, partial model pool size $P$, and evaluation number $S$. We tried setting Tw as 100 and 150 with fixed $P=5$ and $S=900$. The resulted models of these two settings show no significant difference in top-1 accuracy (less than 0.1\%), shown as in Fig.~\ref{fig:hyper}.
Thus we choose warm-up training epochs as 100 in our experiment to save computation resources. For the influence of $P$ and $S$, we show the results in Fig.~\ref{fig:hyper}. It can be seen that the top-1 accuracy of the models found by PC-NAS increases with both P and S. Thus we choose $P= 5$, $S= 900$ in the experiments for better performance. we did not observe significant improvement when further increasing these two hyperparameters.

\textbf{Effectiveness of shrinking search space:} To assess the role of space shrinking, we trains the supergraph of our large space using One-Shot\citep{Brockoneshot} method without any shrinking of the search space. Then we conduct model search on this supergraph by progressively updating a partial model pool in our method. The resulted model using this setting attains $77.1 \%$ top-1 accuracy on ImageNet, which is $1 \%$ lower than our PC-NAS-L as in Table.\ref{ablation}.

We add another comparison as follows. First, we select a batch of models from the candidates of our final pool under small space and evaluate their stand alone top-1 accuracy. Then we use One-Shot to train the supergraph also under small space without shrinking. Finally, we shows the model rankings of PC-NAS and One-Shot using the accuracy obtained from inferring the models in the supergraphs trained with the two methods. The difference is shown in Fig.~\ref{fig:correlation}, the pearson correlation coefficients between stand-alone accuracy and accuracy in supergraph of One-Shot and PC-NAS are 0.11 and 0.92, thus models under PC-NAS's space shrinking can be ranked by their accuracy evaluated on sharing weights much more precisely than One-Shot.

\textbf{Effectiveness of our search method:} 
To investigate the importance of our search method,
we utilize Evolution Algorithm (EA) to search for models with the above supergraph of our large space trained with One-Shot. The top-1 accuracy of discovered model drops furthur to $75.9 \%$ accuracy, which is $2.2 \%$ lower than PC-NAS-L . We implement EA with population size 5, aligned to the value of pool size $P$ in our method, and set the mutation operation as randomly replace the operator in one mixop operator to another. We constrain the total number of validation images in EA the same as ours. The results are shown in Table.\ref{ablation}.
\begin{table}[h]
\caption{Comparision with One-Shot and Evolution Algorithm}
\label{ablation}
\begin{center}
\begin{tabular}{c|c|c}
\hline
\multicolumn{1}{c|}{\bf training method}  &\multicolumn{1}{c|}{\bf search method} &\multicolumn{1}{c}{\bf top-1 acc}\\ 
\hline
Ours      &Ours	   &\textbf{78.1\%} \\
\hline
One-shot  &Ours    &77.1\% \\
One-shot  &EA      &75.9\% \\
\hline
\end{tabular}
\end{center}
\end{table}
\section{Conclusion}
In this paper, a new architecture search approach called PC-NAS is proposed. We study the conventional weight sharing approach from Bayesian point of view and identify a key issue that compromises the effectiveness of shared weights. With the theoretical insight, a practical method is devised to mitigate the issue. Experimental results demonstrate the effectiveness of our method, which achieves state-of-the-art performance on ImageNet, and transfers well to COCO detection and person re-identification too.

\bibliography{iclr2020_conference}
\bibliographystyle{iclr2020_conference}

\appendix
\section{Appendix} \label{section:appendix}

\subsection{Construction of the Search Space:} \label{search_space}
\label{sec:searchspace}
The operators in our spaces have structures described by either Conv1x1-ConvNxM-Conv1x1 or Conv1x1-ConvNxM-ConvMxN-Conv1x1. We define \textbf{expand ratio} as the ratio between the channel numbers of the ConvNxM in the middle and the input of the first Conv1x1.
\paragraph{Small search space} Our small search space contains a set of MBConv operators (mobile inverted bottleneck convolution \citep{sandler2018mobilenetv2}) with different kernel sizes and expand ratios, plus Identity, adding up to 10 operators to form a mixoperator. The 10 operators in our small search space are listed in the left column of Table \ref{operator-table}, where notation OP\textunderscore X\textunderscore Y represents the specific operator OP with expand ratio X and kernel size Y.
\paragraph{Large search space} We add 3 more kinds of operators to the mixoperators of our large search space, namely NConv, DConv, and RConv. We use these 3 operators with different kernel sizes and expand ratios to form 10 operators exclusively for large space, thus the large space contains 20 operators. For large search space, the structure of NConv, DConv are Conv1x1-ConvKxK-Conv1x1 and Conv1x1-ConvKxK-ConvKxK-Conv1x1, and that of RConv is Conv1x1-Conv1xK-ConvKx1-Conv1x1. The kernel sizes and expand ratios of operators exclusively for large space are lised in the right column of Table \ref{operator-table}, where notation OP\textunderscore X\textunderscore Y represents the specific operator OP with expand ratio X and K=Y.

 There are altogether 21 mixoperators in both small and large search spaces. Thus our small search space contains $10^{21}$ models, while the large one contains $20^{21}$.

\begin{table}[h]
\caption{operator table}
\label{operator-table}
\begin{center}
\begin{tabular}{lll|lll}
\hline
\multicolumn{3}{c|}{\bf Operators in both } &\multicolumn{3}{c}{\bf Operators exclusively in} \\
\multicolumn{3}{c|}{\bf large and small space} &\multicolumn{3}{c}{\bf large space} \\
\hline
MBConv\textunderscore1\textunderscore3 & &MBConv\textunderscore3\textunderscore3 &NConv\textunderscore1\textunderscore3 & &NConv\textunderscore2\textunderscore3 \\
MBConv\textunderscore6\textunderscore3 & &MBConv\textunderscore1\textunderscore5 &DConv\textunderscore1\textunderscore3 &  &DConv\textunderscore2\textunderscore3 \\
MBConv\textunderscore3\textunderscore5 & &MBConv\textunderscore6\textunderscore5 &RConv\textunderscore1\textunderscore5 & &RConv\textunderscore2\textunderscore5 \\
MBConv\textunderscore1\textunderscore7 & &MBConv\textunderscore3\textunderscore7 &RConv\textunderscore4\textunderscore5 & &RConv\textunderscore1\textunderscore7 \\
MBConv\textunderscore6\textunderscore7 & &Identity &RConv\textunderscore2\textunderscore7 & &RConv\textunderscore4\textunderscore7 \\
\hline
\end{tabular}
\end{center}
\end{table}
\subsection{Specifications of resulted models:} \label{search_result}
\begin{figure}[htbp]
    \centering
    \includegraphics[width=1\linewidth]{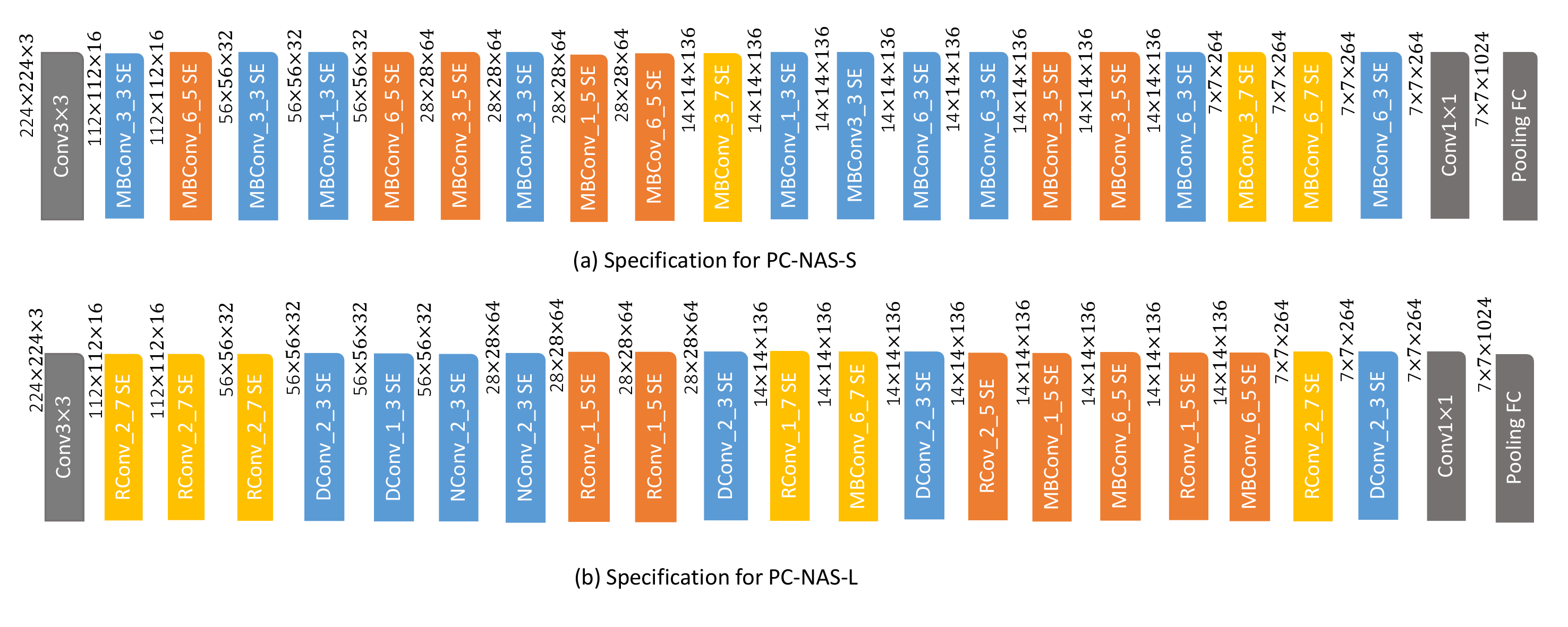}
    \caption{The architectures of PC-NAS-S and PC-NAS-L.}
    \label{fig:BNAS}
\end{figure}

The specifications of PC-NAS-S and PC-NAS-L are shown in Fig.~\ref{fig:BNAS}. 
We observe that PC-NAS-S adopts either high expansion rate or large kernel size at the tail end, which enables a full use of high level features. However, it tends to select small kernels and low expansion rates to ensure the model remains lightweight. PC-NAS-L chooses lots of powerful bottlenecks exclusively contained in the large space to achieve the accuracy boost. The high expansion rate is not quite frequently seen which is to compensate the computation utilized by large kernel size. Both PC-NAS-S and PC-NAS-L tend to use heavy operator when the resolution reduces, circumventing too much information loss in these positions. 

\end{document}